# Real-Time Optimal Trajectory Planning for Autonomous Vehicles and Lap Time Simulation Using Machine Learning

S. Garlick[a] and A. Bradley[b*]

[a]Department of Computer Science,
 The University of Manchester, M13 9PL, UK

[b]Autonomous Driving & Intelligent Transport Group, School of Engineering, Computing & Mathematics, Oxford Brookes University, OX33 1HX, UK

## Abstract

*Widespread development of driverless vehicles has led to the formation of autonomous racing, where technological development is accelerated by the high speeds and competitive environment of motorsport. A particular challenge for an autonomous vehicle is that of identifying a target trajectory – or, in the case of a competition vehicle, the racing line. Many existing approaches to finding the racing line are either not time-optimal solutions, or are computationally expensive - rendering them unsuitable for real-time application using on-board processing hardware. This study describes a machine learning approach to generating an accurate prediction of the racing line in real-time on desktop processing hardware. The proposed algorithm is a feed-forward neural network, trained using a dataset comprising racing lines for a large number of circuits calculated via traditional optimal control lap time simulation. The network predicts the racing line with a mean absolute error of ±0.27m, and just ±0.11m at corner apex - comparable to human drivers, and autonomous vehicle control subsystems. The approach generates predictions within 33ms, making it over 9,000 times faster than traditional methods of finding the optimal trajectory. Results suggest that for certain applications data-driven approaches to find near-optimal racing lines may be favourable to traditional computational methods.*

**Keywords:** Autonomous racing vehicle, Lap time simulation, Optimal racing line, Trajectory planning, Neural network, Machine learning.

---

[*] *Corresponding author. E-mail: abradley@brookes.ac.uk*

# 1 Introduction

Widespread development of Autonomous Vehicles (AVs) has the potential to save many lives, since the overwhelming majority of road accidents are caused by human error (NHTSA, 2015). This has led to the advent of autonomous racing, where fierce competition in motorsport, coupled with the sandboxed environment of the racetrack provides the motivation and ideal testbed to rapidly test, develop and enhance new technology (Skeete, 2019). Here, the vehicle may not have a human driver - indeed, it may not even be possible for a human to drive the vehicle (Betz *et al.*, 2019). The vehicle must be able to map the environment and plan the trajectory itself - thus it is essential for the vehicle to compute the target path which it will then follow.

Lap time simulators (LTS) are commonly employed in motorsport as a cost-effective, efficient tool to evaluate the effect that adjustments in vehicle setup will have upon the resulting lap performance - and run optimisations to find the 'ideal' setup with which to begin real-world testing. Unfortunately, differences in speed profiles, corner radii and other conditions at different circuits result in the 'ideal' setup only being suited to a single track, and thus individual optimisation is required for each circuit.

The majority of LTS in use by race teams are based around Quasi-Steady-State (QSS) approaches (Dal Bianco *et al.*, 2018) - where a vehicle model follows a target path (typically discretised into a series of curves of varying radii) at the highest possible speed, obtained through iteration. This method can be further extended to incorporate transient vehicle behaviour (Völkl *et al.*, 2013). The QSS approach typically requires pre-existing knowledge of a target path, which is usually obtained from logged data of the line taken by a professional racing driver (Colunga and Bradley, 2014) - however, "even with the aid of global positioning systems, the task [of recreating a driven line] is both expensive and rather difficult" (Brayshaw and Harrison, 2005). This assumes the driver took the correct line in the first place, and indeed that it is possible for a human to physically drive the vehicle on track prior to the race. While recreation of a circuit, *e.g.* from aerial photography, may be an option (Salazar-Lozano *et al.*, 2021), for smaller teams - racing on a budget and running design studies with QSS simulators - the racing line may be unreliable, estimated, subject to noise or simply unavailable.

Conversely, for an AV, it is essential for the control system to plan the trajectory the vehicle should take - ideally in real-time, such that the path can be planned without prior processing, thereby enabling the AV to respond to varying road and traffic conditions. To accelerate development in this area, several autonomous racing competitions feature path-planning challenges where the track layout may be unknown (NATCAR, 2021; White, 2020), only available a short time in advance (NXP, 2020), measured using solely rudimentary techniques (FSAE, 2021), or incorporate obstacles to vary the drivable area (FormulaPi, 2018; Zilly *et al.*, 2020; DeepRacer, 2021; F1Tenth, 2021). In such competitions, it may be possible for the vehicle to map the circuit during an initial sighting lap (or learn changes in track layout on-the-fly) - in order to plan a trajectory and build speed on subsequent laps (Kabzan *et al.*, 2020; Kapania *et al.*, 2016). These events pose a significant challenge - the speeds involved in racing demand rapid planning of the trajectory, the impact of vehicle mass upon lap time (Patil *et al.*, 2016) and power requirements dictate achieving this using the minimum possible computational hardware, and the planned trajectory must complete a lap in the minimum possible time.

Traditionally, more rapid approaches to calculating a target path tend to be based purely upon circuit geometry, generating an approximation that "provides no guarantee of finding the time-optimal solution" (Kapania *et al.*, 2016). Many existing algorithms for finding the optimal racing line are based around solving multiple simulations of a vehicle model for a wide range of possible trajectories, before converging upon the path resulting in the minimum time - thus the iterative or optimisation process requires considerable computational resource (Lenzo and Rossi, 2020). This can be time-consuming for traditional LTS, and renders these approaches largely unsuitable for real-time applications.

The ability to expedite the process of generating an accurate trajectory, such that it will run in real-time, would present a significant advantage for an autonomous racing vehicle. Additionally, the ability to rapidly ascertain the ideal racing line for a particular circuit would be beneficial for use in traditional LTS applications, or as a pre-calculation for more accurate and time-consuming optimisations.

In other fields, model-based machine learning (ML) techniques are commonly used to expedite the process of solving a problem by learning from extensive training data to generate extremely rapid predictions of the solution, using a trained Artificial Neural Network (ANN). These approaches have been shown to solve traditionally computationally expensive problems not only quickly, but with a high level of accuracy in the solution (AlQuraishi, 2019). Redmon and Farhadi (2018)'s YOLO object classification algorithm is commonly employed in autonomous racing vehicles, reducing the time taken to process a camera image to a few milliseconds, allowing accurate object identification in real-time at high frame rates - based upon a neural network trained upon several thousand manually labelled images (Culley *et al.,* 2020).

Essentially the ML approach to this type of problem is not one of computationally *solving* the problem, but rather of generating an accurate prediction of the solution to the problem, based upon previously having 'seen' (*i.e.* been trained upon) many solutions to similar problems. While neural networks have been implemented for vehicle control, there have been seldom few attempts to apply the potential speed and accuracy offered by ML techniques to the specific task of generating a target racing line (Capo & Loiacono, 2020).

This paper presents an ANN approach to predicting the ideal racing line, aimed at reducing the calculation time by several orders of magnitude. The network can be trained on data from any existing method of optimal trajectory generation, thus facilitating predictions based upon highly complex models to be made within milliseconds. This enables the target path to be calculated much more rapidly than with traditional methods - reducing computational burden for traditional lap time simulators, and facilitating real-time application in an autonomous racing vehicle using on-board hardware. This original approach provides an unprecedented reduction in the time taken to generate the ideal racing line, with minimal loss in accuracy of the solution.

## 1.1 Related work

Simple approaches to finding a target path are commonly based around following the shortest path or the Minimum Curvature Path (MCP), which Heilmeier *et al.* (2019) were able to calculate in around 18 seconds for the *Roborace* autonomous racing car. Whilst the MCP can be calculated rapidly, it merely offers an approximation of a racing line - which may not result in a fast lap (Kapania *et al.*, 2016) - and includes no consideration of the peculiarities of different vehicles. Where a higher degree of accuracy in the resulting racing line (thus the resulting lap time and consequent vehicle setup) is required, this approach is insufficient.

The majority of existing methods of calculating a time-optimal solution are based around the free-trajectory, Optimal Control Problem (OCP) approach to lap time simulation - where the simulator generates the driver commands for a dynamic model of the vehicle to follow any path within the circuit bounds. The vehicle throttle, brake and steer inputs are typically optimised at points on a discretised circuit such that the resulting velocity and trajectory result in the minimum time (Casanova, 2000) - thus delivering the optimal racing line. This approach enables complex, fully transient models of the vehicle to be used (Kelly, 2008), and allows the path to be adjusted in response to a change in vehicle parameters (Dal Bianco *et al.*, 2018) - resulting in a highly-accurate, vehicle-specific racing line. However, due to the complexity of solving a trajectory of the vehicle model for every discretisation point, the solution time for the OCP approach is typically considerably longer than the lap time (Lenzo and Rossi, 2020) - with simulators comprising relatively simplistic vehicle models taking around 15 minutes to solve a lap (Brayshaw and Harrison, 2005; Perantoni and Limebeer, 2014). Lot and Dal Bianco (2015)'s more complex model with 14 Degrees of Freedom solved a lap in approximately 28 minutes. The free-trajectory approach (required to find the racing line) typically takes considerably longer to solve than a LTS which follows a predetermined path (Veneri and Massaro, 2019). While OCP approaches offer accurate, time-optimal solutions to the racing line, the computation times typically render them unsuitable for use in real-time applications.

In an AV, it is necessary to perform these calculations in real-time or faster, using on-board processing hardware. Jain and Moraro (2020) addressed this using Bayesian optimisation, resulting in calculation of a racing line in under three minutes. Though this enables rapid pre-computation of a new circuit after an initial sighting lap, it is not fast enough to calculate the subsequent lap *whilst* driving the previous at speed. Christ *et al.* (2019) modelled varying tyre-road friction around the Berlin *Formula E* circuit for the autonomous *Roborace* vehicle, using CasADi (Andersson *et al.*, 2019) to reduce simulation time to under two minutes - again slower than the lap time. Kapania *et al.* (2016) achieve quasi real-time path planning using an iterative, two-step process (using the MCP as a pre-calculation) to generate a trajectory for an autonomous racing car within 26 seconds - *i.e.* faster than the lap time. However, the vehicle behaviour has been simplified to a bicycle model - resulting in an approximation of the racing line which (though generally accurate), exhibits areas where "a significant discrepancy on the order of several meters between the two-step algorithm's trajectory and the other comparison trajectories" exists (Kapania *et al.*, 2016).

The use of machine learning has previously been explored for real-time vehicle control - commonly arriving at an optimal path as a by-product of the resulting vehicle motion. For example, Salem *et al.* (2019) coupled evolutionary learning with a fuzzy-logic controller to refine the vehicle control inputs and racing line, resulting in a controller which can drive the vehicle around a circuit based upon learnings from previous iterative experimentation – noting that further work is required to drive more than a single circuit. Yu *et al.* (2016) used Reinforcement Learning (RL) to learn simple control inputs for a basic 1980s *Outrun* racing game (Suzuki, 1986), while Balaji *et al.* (2020) developed an end-to-end RL approach in a simulator to learn to drive a real (1/18th scale) vehicle from camera imagery - though the vehicle is incentivised to follow the track centreline, rather than a racing trajectory.

An alternate ML approach is to learn from a human driver. Fridman *et al.* (2019) trained an end-to-end autonomous driving algorithm using 4.2 million video frames acquired from a *Tesla* - however, the majority of approaches for roadgoing applications are aimed at safely navigating the road environment, rather than generating a time-optimal path. An end-to-end approach was trialled for autonomous racing by Koppula (2017), noting that the method was "insufficient to produce smooth steering behaviors" in simulation. In any case, end-to-end approaches typically generate the vehicle control inputs from sensor or camera footage (Tampuu *et al.*, 2020) rather than a target path from the road ahead.

Few studies exist which directly use ML for the specific task of generating a racing line. Cardamone (2010) employed a genetic evolution ML technique to search for the ideal line by iteratively refining the target path, obtaining the optimal trajectory for a specific circuit. The simple vehicle model used limits the accuracy of the generated line, although it offers a considerable improvement over the MCP. A similar technique was employed by Vesel (2015) who added a 'healing' sub-process to improve performance - though these approaches are only able to provide a trajectory for the specific circuit they trained upon. Weiss & Behl (2020) employed a CNN to learn a racing trajectory ahead of the vehicle in a racing game, subsequently generating vehicle control inputs and outperforming traditional end-to-end methods – though it is unclear whether the system is able to drive a previously-unseen circuit. To this end, Capo & Loiacono (2020) used reinforcement learning to plan a short-term trajectory directly ahead of the vehicle for gaming applications. While the method attempts to minimise lap time, *and* is able to plan a trajectory for a previously-unseen circuit, the "learned behaviour is still far from the performance of professional racing drivers" (Capo & Loiacono 2020) - due in part to the description of the target path as a single point ahead of the vehicle.

Currently, there is no widespread conventional method of generating a highly accurate racing line for a previously-unseen circuit in real-time or faster. ML techniques offer the potential to significantly reduce solution time, and thus their application to the task of finding the ideal racing line offers the tantalising potential of being able to plan a racing line extremely rapidly with minimal loss of accuracy - thus facilitating use in the real-time application of an autonomous racing vehicle.

# 2 Methodology

This study employs a feed-forward ANN, trained upon a dataset of racing lines for a variety of circuits. This enables racing lines to be rapidly generated for previously-unseen circuits, based upon trajectories found in the training data. Typically, for an ANN to make accurate predictions, it must have been trained on a dataset that encompasses the entire range of features which may be found in the new, unseen problem – and therefore an extensive dataset was generated. Inner and outer boundaries were reconstructed for a large number of real circuits, which were then augmented to expand the dataset to contain over 6,000 tracks and split into 2.7 million individual segments. An optimal racing line was generated for each circuit using an existing OCP approach to lap time simulation, thus completing the dataset.

Each circuit is split into smaller segments using the sliding window approach - enabling the network to learn the racing line for multiple sections of the track rather than an entire circuit. This enables the network to operate on circuits of any length, and generalise track sections across multiple circuits. Once segmented, data describing the vehicle's position on track and the surrounding circuit geometry were extracted for various waypoints around the lap, and these 'features' were used to train the network using the majority (86%) of the dataset.

Once trained, the ANN can calculate the features for a new, previously-unseen circuit - thus generating a prediction of the racing line. The hyperparameters of the network were tuned to improve the accuracy of the prediction, using a further 9% of the dataset. K-Fold analysis (Stone, 1974) was used for this purpose.

Finally, the overall performance of the system is quantified by comparing the generated racing lines against the remaining 5% of the dataset - comprising circuits which have not previously been seen by the network. The methodology is described in the following subsections, with results presented in section 3.

## 2.1 Training data generation

The dataset used in this study contains inner and outer track boundaries, and the ideal racing line for each circuit. There is no restriction placed upon the method used to generate the optimal line used in the dataset - it could be taken from a detailed, high-order free-trajectory OCP method (*e.g.* Lot and Dal Bianco (2015)), a human in a Driver-in-Loop simulator - in which case it would 'learn' the lines preferred by a particular driver) - or any other traditional method of acquiring the racing line.

Prior to generation of the optimal line in the training data, it was first necessary to recreate a large number of circuits. Inner and outer boundaries were obtained for 82 real circuits from around the world, including several tracks from the Formula 1 calendar (60 manually generated, 22 taken directly from TUMFTM (2020)). As is common practice in ML studies (Culley *et al.,* 2020), augmentation was used to expand the dataset to a total of 6058 circuits by scaling, flipping, and reversing the direction of travel for each track. Care was taken to ensure that an equal proportion of real circuits were in the training and validation data, with a larger proportion of real circuits retained for final testing. 288 circuits (comprising 9 real-world, and, and 279 augmentations thereof) were used as testing data to evaluate the performance of the network on a completely new, unseen circuit.

To generate the optimal trajectory for each circuit in the dataset, the *Global Race Trajectory Optimisation* tool (TUMFTM, 2019) developed by Christ *et al.* (2019) was selected, using the default solver settings and vehicle parameters for the *Roborace* vehicle. This consists of a double-track vehicle model with QSS weight transfer and non-linear tyre models, converted to an NLP using Gauss-Legendre collocation, and solved using IPOPT – with reduced computation times achieved through a curvilinear abscissa track description and use of the CasADi framework. This provides a balance between the accuracy of the generated line, and the time taken to calculate optimal paths for each of the individual circuits in the dataset. The time-consuming process of solving over 6000 optimal laps would have taken approximately two weeks on a single processor thread – however, to expedite the task the workload was split across multiple PCs with multi-core processors.

Further expansion of the training data was achieved by splitting each circuit into a series of overlapping segments (or windows), meaning that the entire dataset contains over 2.7 million segments of different racetracks - complete with optimal racing lines for each segment. The rationale, process and advantages of this segmentation are outlined in section 2.2. The total size of the dataset required for accurate prediction of the ideal racing line was established experimentally, based upon a similar methodology to Roh *et al.* (2019).

All optimal trajectories were calculated assuming a vehicle width of zero – *i.e.* the vehicle centre uses the full width of the track. This enables removal of the vehicle width parameter from the training data - instead accounting for it when the network predicts the target path, thus facilitating adjustment of the vehicle width after the network has been trained.

The complete dataset comprising inner and outer circuit boundaries, and optimal racing lines was divided into subsets in accordance with Table 1.

| Dataset | Real | Augmented | Total |
| --- | --- | --- | --- |
| Training Data | 66 | 5168 | 5234 |
| Validation Data | 7 | 529 | 536 |
| Testing Data (Unseen) | 9 | 279 | 288 |
| **Total** | **82** | **5976** | **6058** |

*Table 1*. *Number of circuits contained in each subset of the dataset.*

## 2.2 Circuit segmentation via sliding windows

Each circuit was divided with a series of lines (or Normals) perpendicular to the track centreline, and the position upon these lines at which the vehicle trajectory intersects was identified as a 'waypoint'. Whilst it is common (Botta *et al.*, 2012; Casanova, 2000) to vary the spacing between the lines segmenting the circuit to reduce resolution on straights (thereby reducing computational effort), a fixed interval of five metres was used. Maintaining a constant interval is beneficial for the network as it eliminates an additional variable, thus reducing dimensionality of the input. In the event that two or more Normal lines intersect one another (*e.g.* at a tight hairpin bend), the lines are modified slightly to become 'Pseudo-Normals' - where the angle between the Normal and the track centre line is adjusted away from 90 degrees until the lines no longer intersect (figure 1). This is encoded and fed into the network in section 2.3.

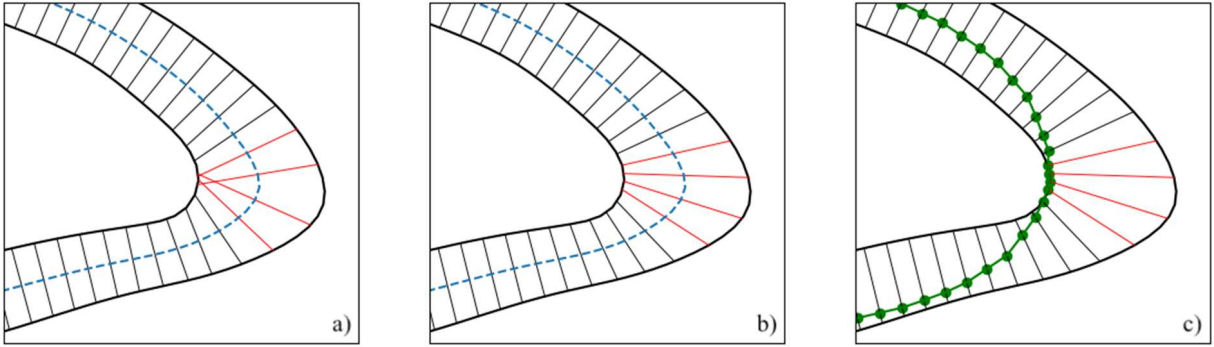

*Figure 1.* - a) Lines normal to the track centreline divide the circuit, b) Normals adjusted to become 'pseudo-Normals' at tight hairpins, c) Waypoint location identifies the intersection of the racing line.

In the same way that a human driver looks ahead on the circuit, several other studies (Casanova, 2000; Weiss and Behl, 2020) use a forward-looking 'preview' to control the subsequent behaviour of the vehicle. However, the racing line at any point on the circuit is influenced by the preceding curvature (thus resulting in the current vehicle position) and the upcoming curvature of the circuit. Therefore, a performance improvement was found by introducing a bidirectional 'Foresight' - both fore and aft of the current position - enabling the network to learn both how the vehicle arrived at the current position, and where the trajectory heads next.

Rather than simply passing all the track Normals for each entire circuit to the neural network, they are broken into sections using the sliding window approach. Thus, the circuit is described as a series of overlapping windows, each comprising a fixed number of Normal lines and subsequent waypoints. This approach has the following benefits; the network input is a fixed size, and therefore is capable of handling circuits of different length (*e.g.* Nordschleife (20.8km) and Brands Hatch GP (2.9km)); the network learns from a larger number of windows of racetracks, enabling it to generalise across behaviour from multiple circuits with similar features; the sliding window approach vastly increases the quantity of data provided by each circuit, thereby reducing the number of circuits required in the training dataset.

To represent this mathematically, each circuit is represented by a number of overlapping windows, containing information for each of the *f* Normal lines fore and aft of the central Normal within the window. For a given Foresight *f*, the window ($t_i$) is formed such that it contains the ordered set of track Normals surrounding and including the Normal line $N_i$ ($N_{i-f}$, ..., $N_i$, ..., $N_{i+f}$). The neural network *g* is then trained upon these windows in order to predict where the waypoints lie upon a given track section, $g(t_i) \rightarrow w_i$, where $w_i$ is the waypoint location on the Normal $N_i$. Tuning of the Foresight is explained in section 2.4.2. The trajectory for an unseen track is later calculated by predicting the waypoints for each window of the new circuit using the fully trained network.

In this study, the Foresight is symmetric and tuned for the objective of minimising the prediction error for a circuit given to the network in its entirety. However, the windowed approach means that the trajectory is being repeatedly planned for the area immediately surrounding the vehicle - and thus the Foresight could theoretically be linked to that of an AV's perception sensors, making this technique suitable for rapid real-time trajectory planning in an autonomous road-going vehicle.

## 2.3 Feature extraction

Rather than learning the racing line for a complete circuit, the network is trained upon information about how the circuit and corresponding racing line flow through each of the windows described in section 2.2. This provides the ANN with sufficient data to infer important semantic information, in order to generalise a function that creates a target trajectory. It is necessary to encode this information about the circuit and racing line in a lossless manner. Therefore, features are extracted from the training data including circuit width (or length of each Normal), the racing line position on track (or waypoint position upon the Normal), and information describing the curvature of the circuit at each Normal.

Many existing LTS describe the target racing line as a curvature profile, facilitating calculation of maximum cornering velocities (Colunga and Bradley, 2014; Dal Bianco *et al.*, 2019). However, as the circuit twists and turns, the radius of curvature varies in a highly nonlinear manner, reducing to a few metres at hairpins, and tending towards infinity on straights. To constrain these feature representations, the circuit is represented by variations between the Normal lines describing the structure of the circuit, rather than the curvature (figure 2).

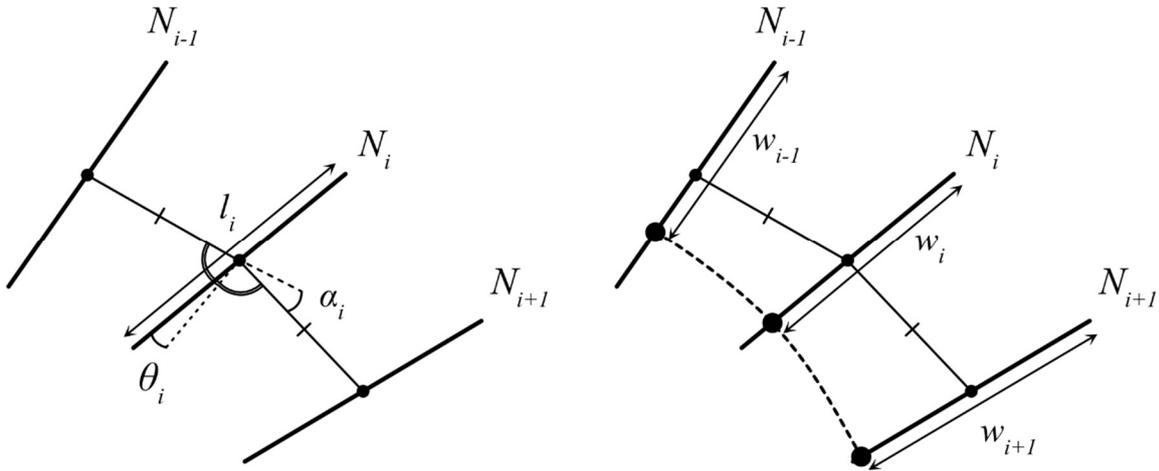

***Figure 2.*** *Feature extraction: Track geometry (left) and waypoints describing the racing line (right).*

The circuit geometry is described by encoding the features for each Normal as follows; length of the Normal *l,* angular change between Normals *α*, and the angle *θ* between the Normal and the true normal to the track centreline (*i.e.* 90 degrees, except in the case of an adjusted 'pseudo-Normal'). These features are encoded into the network by allocating three dimensions in the input space for each Normal within a given window (figure 3). Preserving the sign of each angle enables encoding of both left and right-hand bends.

The racing line is described by identifying the location of the waypoint along each Normal - denoted using the feature *w*, ranging between zero and one representing the left and right of the Normal respectively. The network is then trained upon the *w* values for each window in the training data, thus enabling prediction of a racing trajectory through a new circuit.

## 2.4 Network

The neural network selected for this study is a feed-forward ANN due the nature of the problem being a regression task (Lathuilière *et al.*, 2019), and the minimal computation time required by this type of network in generating a prediction. This section outlines the network structure, and the tuning used to identify the optimal parameters, implemented using the *Keras* framework (Chollet & others, 2015).

### 2.4.1 Network structure

The network structure contains four fully connected layers, with the Huber loss function (Huber, 2004) and Nadam optimiser (Dozat, 2016) - selected following experimentation with the various hyperparameters of the network. The number of units within each layer was chosen using a grid search method - resulting in 450, 200 and 200 units on the first, second and third hidden layers respectively (figure 3).

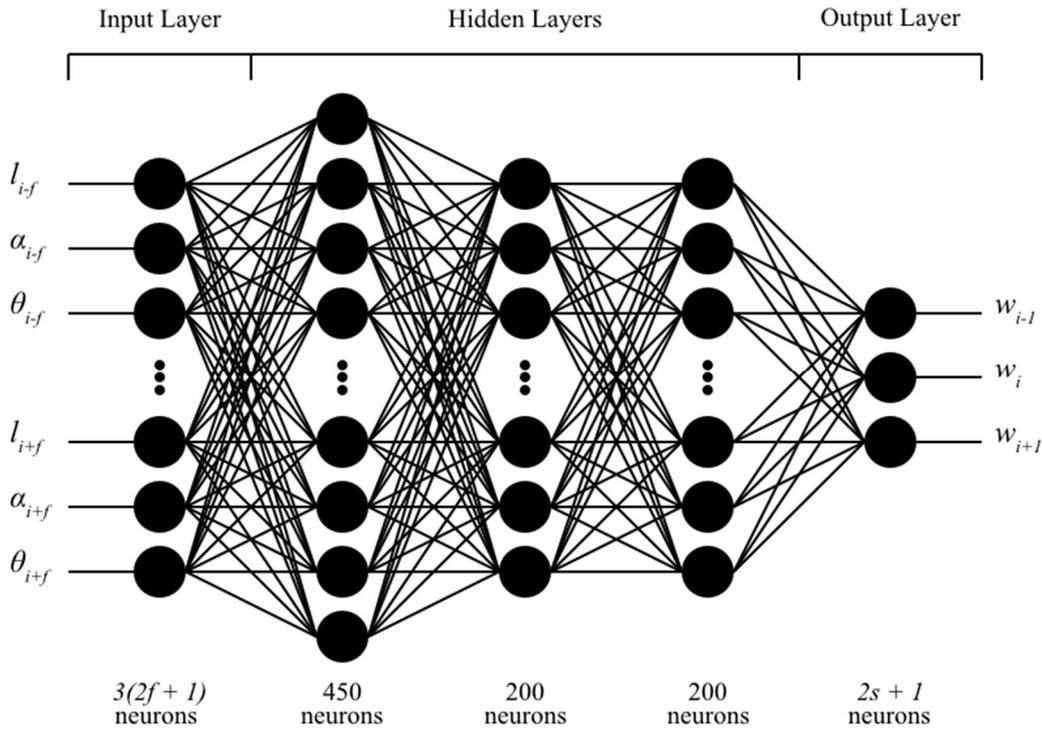

***Figure 3.*** *Network structure (shown with sampling level s=1).*

The network was found to perform best when activation functions of sigmoid and hard sigmoid (Gulcehre *et al.*, 2016) were used across the hidden and output layers respectively. The output dimensions are determined by the sampling level *s* (section 2.4.3) as shown in figure 3. Network training took approximately 30 minutes on a 2.4 GHz Intel Core i9-9980HK processor.

*2.4.2 Foresight*

The Foresight dictates the number of Normals considered within each window. Greater visibility of the previous and upcoming circuit geometry results in an improved prediction of the racing line. However, as Foresight is increased, the ANN's input size (and thus problem complexity) increases (Bauer & Kohley, 2019). In addition, consideration of ever more distant areas of the track becomes increasingly less valuable in determining the optimal line for the current point. Therefore, the optimal value of Foresight occurs where sufficient information about the previous and upcoming geometry is present, without adding excessive dimensionality to the network.

Ideally, the Foresight should be sufficient to cover the length of the longest straight in the dataset - otherwise, in a straight section (where there is no apparent curvature in the circuit) the network has no concept of whether the previous and upcoming bends are left or right-handed, and thus is unable to infer the correct line for this section. The prediction then tends towards the centreline of the straight, resulting in larger errors.

The optimal Foresight is therefore dependent upon the types of circuit contained within the dataset. For the dataset used in this study, 70 waypoints (*i.e.* 350m fore and aft of the central Normal in the window) was found to provide a balance between track coverage and problem complexity (figure 4).

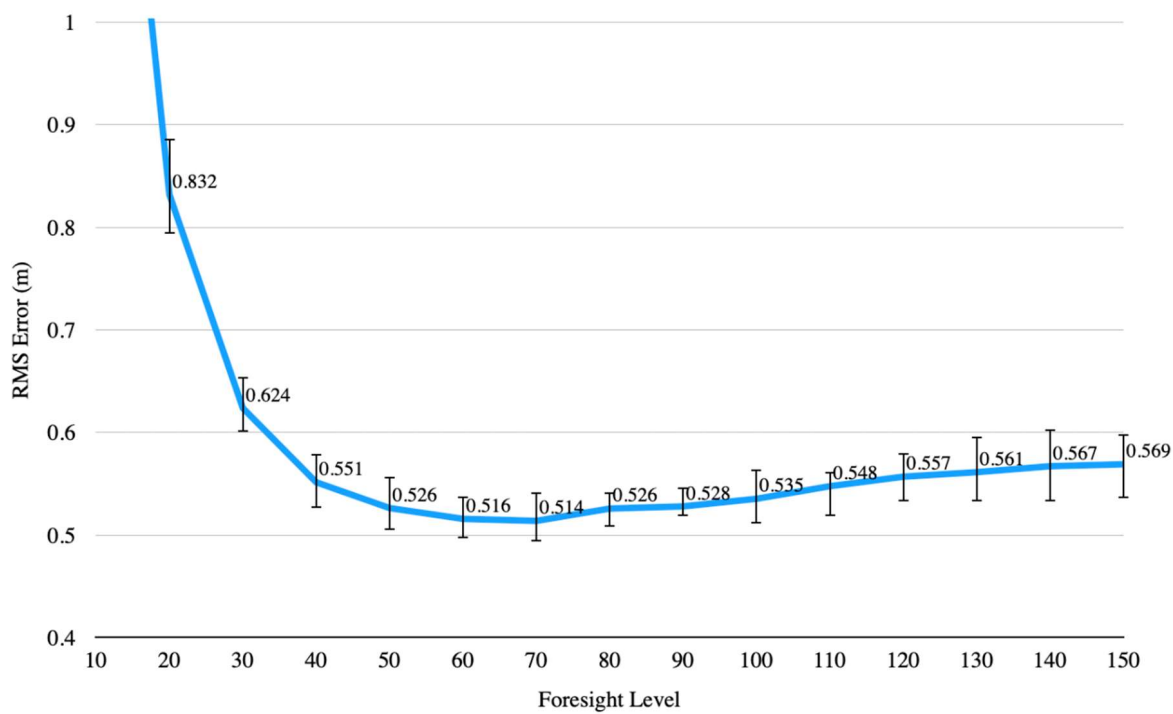

*Figure 4. Effect of Foresight upon racing line prediction accuracy (based upon a subset of the validation dataset). Error bars denote the variance through the different K-Folds.*

*2.4.3 Output sampling*

Minimising the output size of the network reduces dimensionality - improving the ability to generalise, and thus increasing prediction accuracy. Therefore, rather than predicting every waypoint within a window, a smaller number of waypoints are predicted and assembled with those sampled from the neighbouring windows - forming the predicted racing line for the complete lap.

In order to create a racing line which flows smoothly from one waypoint to the next, the network predicts multiple waypoints either side of the central Normal within each window and averages across those sampled from the neighbouring windows (rather than assembling the single, central waypoint within each window, which results in a slightly noisy trajectory). This provides a 'moving average' effect to each predicted waypoint, thereby generating a smooth racing line.

The sampling level (*s*) determines the number of pairs of surrounding waypoints that the network predicts, such that *s = 0* (*i.e.* no averaging) dictates a network with one output and *s = 2* requires a network with five outputs. Once each track segment has been passed through the network, *2s+1* predictions of the waypoint location exist for each Normal – which are averaged to result in a smooth line.

A sampling level of *s=4* (nine network outputs), was found to provide the highest accuracy in the predicted line for the validation data used in this study (figure 5), generating a prediction of the racing line which flows smoothly from waypoint to waypoint without requiring computationally expensive filtering techniques of the nature employed by Heilmeier *et al.* (2019).

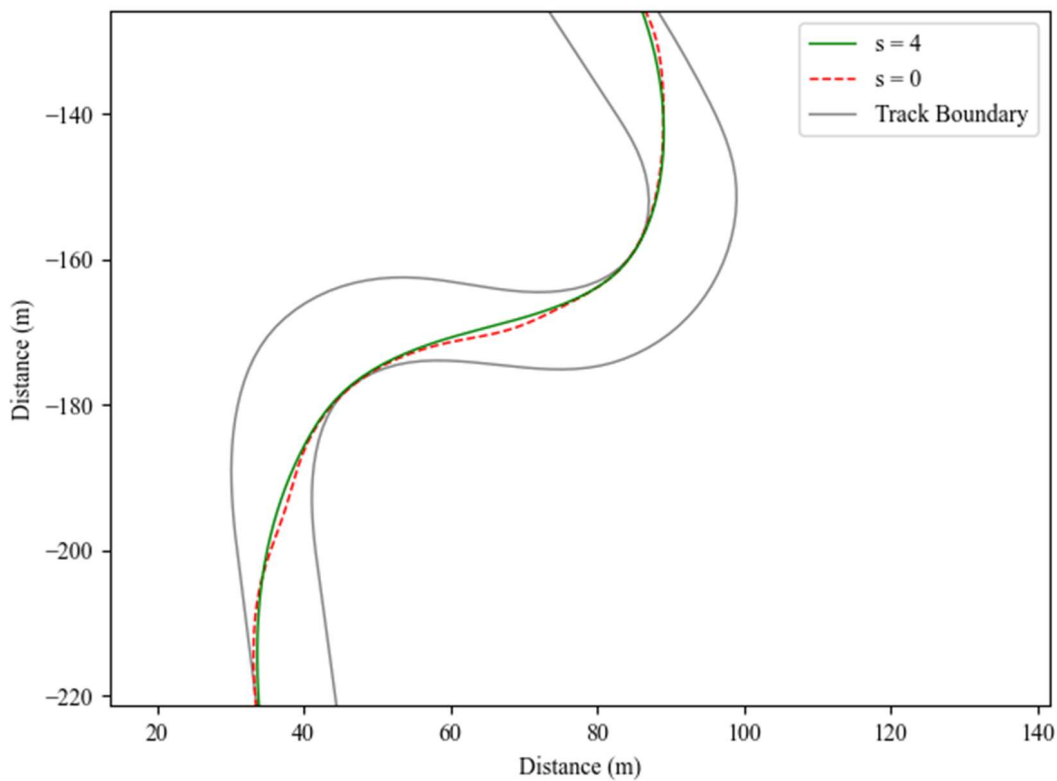

***Figure 5.*** *Adjustment of the Sampling level provides a smoothing effect to the predicted line.*

# 3 Results

To demonstrate the performance of the ANN, predictions were made for the previously unseen circuits in the testing data and compared against the racing lines found using the OCP method provided by TUMFTM (2019). Comparisons were made for both accuracy of the generated line, and solution time.

## 3.1 Racing line prediction accuracy

The racing lines calculated by the ANN and the OCP method for Brands Hatch GP (figure 6) and Nürburgring GP (figure 8) circuits are presented below, with lateral deviation plots to aid comparison (figure 7 & 9).

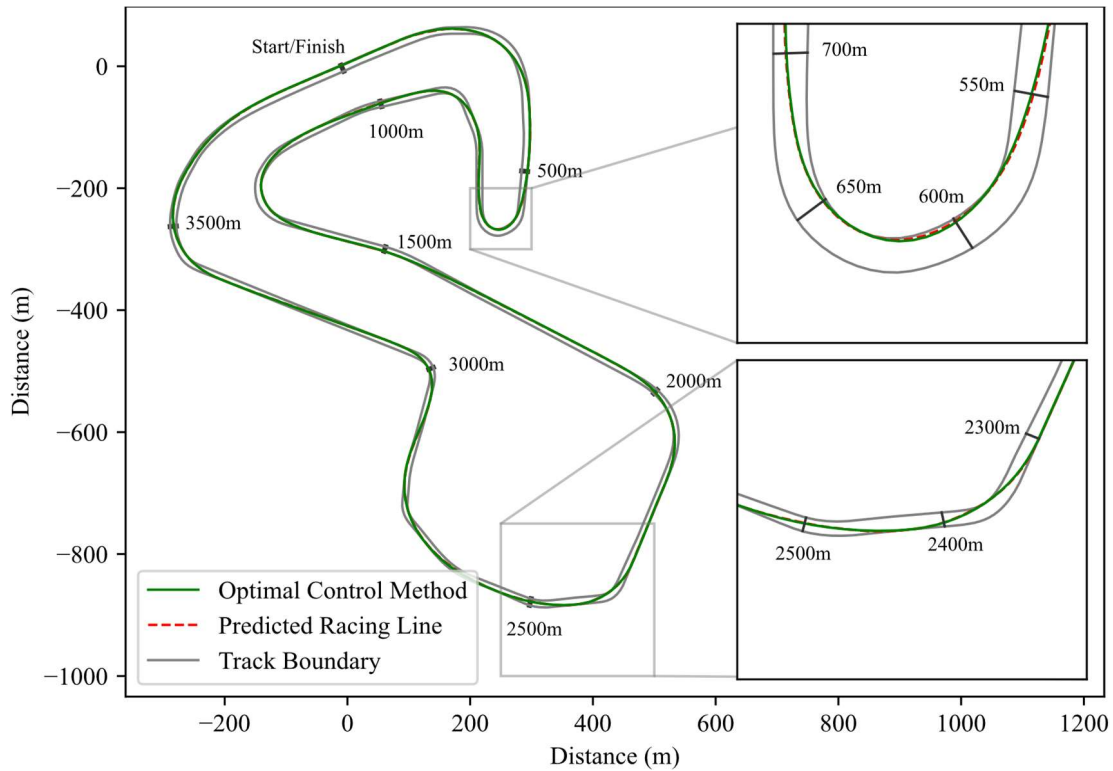

*Figure 6. Comparison of the racing line predicted by the ANN against that calculated via the OCP method for the previously unseen Brands Hatch GP circuit.*

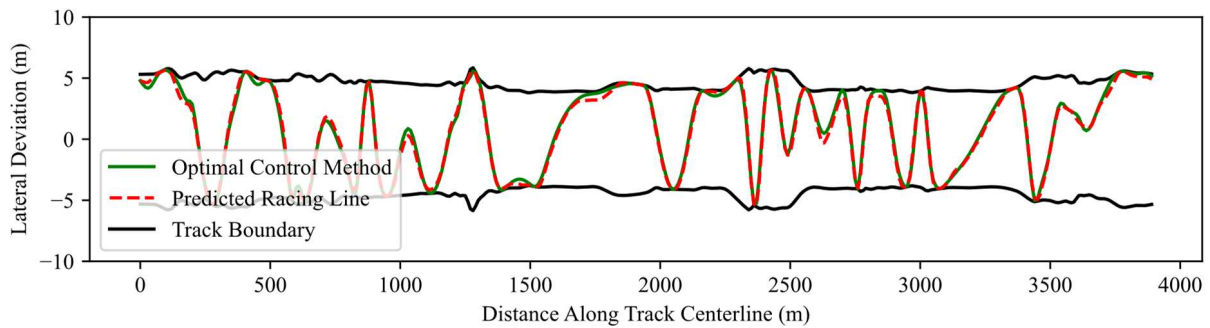

*Figure 7. Lateral deviations from the track centreline predicted by the ANN and calculated via the OCP method for the previously unseen Brands Hatch GP circuit.*

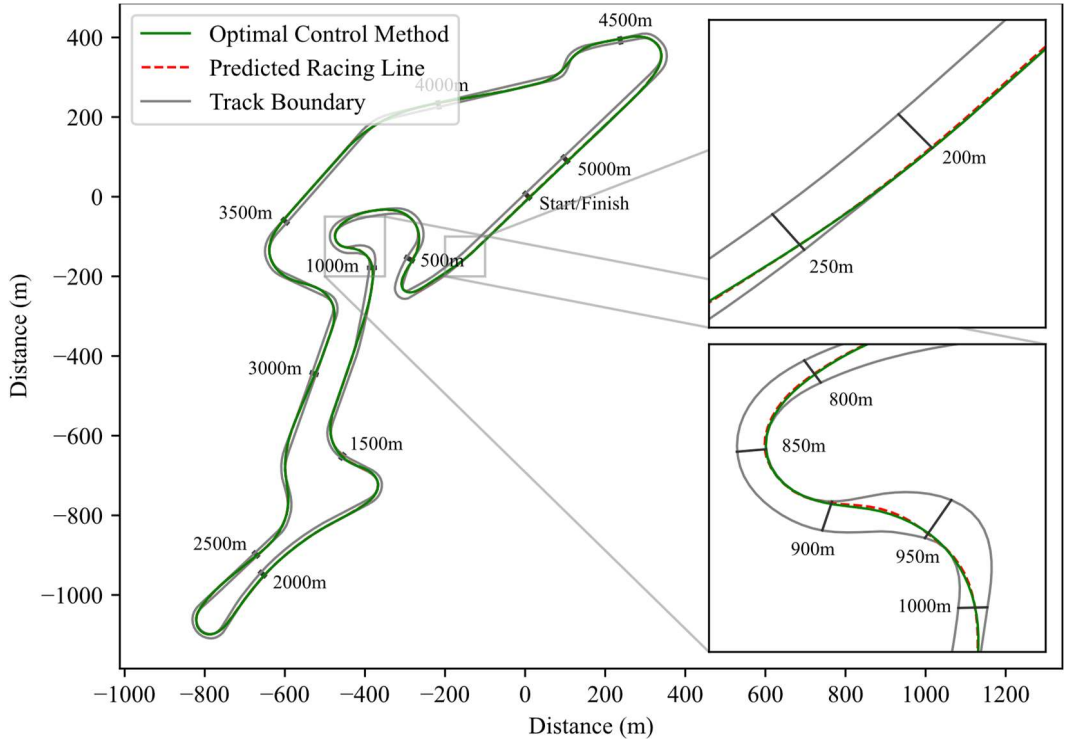

*Figure 8. Comparison of the racing line predicted by the ANN against that calculated via the OCP method for the previously unseen Nürburgring GP circuit.*

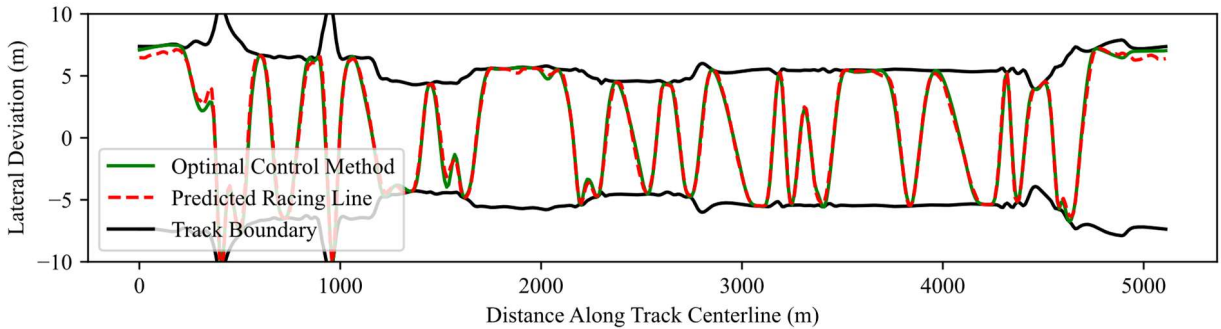

*Figure 9. Lateral deviations from the track centreline predicted by the ANN and calculated via the OCP method for the previously unseen Nürburgring GP circuit.*

The trajectories predicted by the ANN are qualitatively similar to those calculated using the OCP method, following near-identical paths at most apexes - where it has the most impact upon lap time (Kelly, 2008) - owing to the relative consistency of the optimal mid-corner racing line in the training data. Maximum deviations between the ANN's prediction and the OCP method tend to occur in areas with more complex and unusual features - *e.g.* the straight with a subtle bend prior to the upcoming corner at Nürburgring GP, and the complex combination of consecutive turns impacting upon each other (figure 8). Larger sustained deviations tend to occur during straights - however, this behaviour is common to many methods of calculating the optimal line and is considered to be of lesser importance as a difference in vehicle position during a straight results in minimal impact upon lap time (Dal Bianco *et al.,* 2019; Kapania *et al.*, 2016).

Quantitative analysis of the performance for the previously unseen real-world circuits in the testing data is provided in Table 2, demonstrating that the network is capable of generating accurate racing lines for a variety of different track lengths and layouts. The highest accuracy is achieved for circuits such as Brands

Hatch GP and Hungaroring, which comprise features more commonly found in the training dataset such as constant radius turns at the end of a straight, moderate length straights, and average track width.

Circuits further from the 'average' in the training dataset in some way (*e.g.* those which contain complex corners of varying radius, quick successions of multiple bends, very short straights *etc.*) which cause more variance in the racing line may present 'edge cases' - features sparse in the training data, or the ANN has not previously seen. In these cases, the predicted line becomes a more generic approximation biased towards the centre of the dataset, and prediction accuracy reduces. This is demonstrated by the increased prediction error at Paul Ricard (featuring long straights, and double-apex corners of varying radius), and the 20.8km Nordschleife (with an abundance of complex corners, and an extremely long straight). To reduce the error induced by this sample biasing, the dataset should be centred around similar circuits and diversified. Therefore, the average circuit in the training data should contain features similar to those for which predictions are required (*i.e.* if a prediction is desired for a small kart track, training the ANN on a dataset featuring only F1-style circuits would result in reduced accuracy) - and diversification should be achieved through the use of a greater selection of original circuits in the training data (rather than the large number of augmentations of a small number of circuits used in this study).

|  | Prediction accuracy | | | Solution Times | |
| --- | --- | --- | --- | --- | --- |
| **Circuit** | **RMSE (m)** | **MAE (m)** | **Apex Error (m)** | **OCP (s)** | **ANN (s)** |
| **Hungaroring** | 0.331 | 0.255 | 0.112 | 117 | 0.0293 |
| **Nürburgring GP** | 0.354 | 0.266 | 0.160 | 554 | 0.0314 |
| **Nordschleife** | 0.470 | 0.336 | 0.127 | 2670 | 0.0550 |
| **Paul Ricard (1C-V2)** | 0.547 | 0.411 | 0.198 | 468 | 0.0314 |
| **Red Bull Ring GP** | 0.393 | 0.276 | 0.110 | 134 | 0.0288 |
| **Spa** | 0.358 | 0.256 | 0.089 | 208 | 0.0335 |
| **Brands Hatch GP** | 0.285 | 0.226 | 0.105 | 244 | 0.0285 |
| **Monza** | 0.377 | 0.272 | 0.161 | 207 | 0.0327 |
| **Catalunya GP** | 0.398 | 0.302 | 0.116 | 202 | 0.0293 |
| **Average (circuits above)** | **0.390** | **0.289** | **0.131** | **534** | **0.0333** |

*Table 2. Accuracy of the racing line predicted by the ANN in comparison to the OCP method, with respective processing times for the 9 real-word circuits in the testing dataset.*

A quantitative analysis of the prediction error observed across the entire testing dataset is presented in Table 3, delivering a Mean Absolute Error (MAE) of ±0.27m (±0.38m RMSE) averaged across all circuits in the testing dataset - closely echoing the behaviour across the nine real-world circuits. The distribution of errors closely approximates a Laplace distribution (giving a 50% Confidence Interval (CI) of just ±0.18m), with a mean very close to zero. It is the authors' belief that flipping and reversing each circuit in the training data (and thus training the ANN on an equal number of left and right-hand bends) has resulted in this symmetric distribution with (almost) zero offset. At corner apex - where it matters most - the prediction error reduces to just ±0.11m average error across all circuits in the testing dataset.

| RMS (m) | MAE (m) | Mean error (m) | Apex Error (m) | 50% CI (m) | 95% CI (m) |
|---|---|---|---|---|---|
| 0.376 | 0.267 | 0.029 | 0.111m | 0.184 | 0.826 |

*Table 3. Overall accuracy of the ANN predictions in comparison to the OCP method, across the entire testing dataset (comprising 288 circuits).*

The error distribution provides a 95% Confidence Interval (CI) of ±0.83m - while Brayshaw and Harrison (2005) found that lines taken by four professional racing drivers varied by more than ±1m. This implies that the ANN is capable of generating a trajectory which is comparable to, or indeed more accurate than a human racing driver's ability to identify and follow the optimal path - though there are many other factors which may affect a human driver's choice of line. For additional real-world context, the prediction errors are smaller than the ±0.31m MAE change in optimal line due to localised (±10%) variations in friction coefficient on the track surface observed by Christ *et al.* (2019) or the ±1m (95% CI) change in optimal path due to adjustments in vehicle design parameters (Brayshaw and Harrison, 2005).

The accuracy with which the ANN can predict the optimal path is comparable to, or better than, variance between other traditional approaches of calculating an optimal trajectory. Using a qualitative comparison of the racing line and lateral deviation plots presented in this work (figures 6, 7, 8 & 9) with those presented in other LTS studies, the prediction accuracy appears to compare favourably the displacement between Siegler *et al.* (2000)'s QSS and transient calculations of the racing line, and similarly to the discrepancy between Dal Bianco *et al.* (2019)'s direct and indirect methods – where an approximate apex error of ±0.08m can be inferred (*vs* ±0.11m in this study). Kapania *et al.* (2016)'s discrepancy of "several meters" in calculation of the optimal trajectory is significantly greater than the ±0.83m (95% CI) observed in this study. However, it should be noted that several existing methods calculate the trajectory alongside a velocity trace and often vehicle control inputs - which are currently beyond the scope of this study.

From the perspective of AV control, the ANN's prediction accuracy is comparable to Andresen *et al.* (2020)'s accuracy of map generation in an autonomous racing vehicle at racing speed (±0.39m *vs* ±0.29m RMSE respectively). The accuracy is qualitatively similar to that achievable by a path-following driver model following a racing line at speed in an autonomous racing car (Culley *et al.*, 2020; Kapania, 2016), and a road-going AV in Wang *et al.* (2016).

## 3.2 Computation time

Typical solution time for generating a racing line for a previously unseen track via the neural network is 33 milliseconds on a 2.4 GHz Intel Core i9-9980HK processor[2] - over 9,000 times faster than generating the line (for the majority of normal length circuits) via the OCP method. An interesting feature of the ANN approach is that the windows can be calculated simultaneously, while conventional approaches iteratively proceed around the circuit. Thus, solution times for a very long circuit such as the 20.8km Nordschleife (50ms *vs* 46 minutes for the OCP method) are barely increased from that of a shorter track - rendering the ANN approximately 50,000 times faster in generating a racing line for this particular circuit.

The method proposed in this study is approximately 800 times faster than Kapania *et al.* (2016)'s two-step approach - which (to the authors' knowledge) was previously the fastest known method of generating a non-trivial racing line - whilst delivering a considerably more accurate solution.

---
[2] The ANN is ideally suited to GPU/TPU processing - thus performance could be further expedited on compatible hardware.

# 4 Discussion

The qualitative and quantitative comparisons presented in section 3.1 demonstrate that this radically different approach to finding the racing line results in accurate predictions which offer comparable or reduced error to; other traditional approaches for calculating the optimal trajectory; the ability of a human racing driver to identify and follow an optimal path; the effect of small changes in friction or vehicle parameters; an autonomous vehicle's ability to map and track a target path. However, the real benefit of this approach to racing line generation lies not in the accuracy of the prediction, but in the speed and computational efficiency of obtaining an accurate solution.

The ANN delivers an incredibly rapid prediction of the racing line, which represents a vast reduction in solution time over traditional solutions - with minimal sacrifice in accuracy. The current implementation is able to make a prediction for a pre-defined vehicle upon an unknown circuit. However, it does not consider differences in track or weather conditions, or adjustments to vehicle parameters - for which a more extensive dataset including such effects would be required. Since changes to racing lines resulting from differences in vehicle parameters are typically small (Brayshaw & Harrison, 2005), generation of a generic racing line for use in QSS simulations (*i.e.* the majority of widely-used LTS software) is possible using this method.

The technique is ideally suited for generation of the target path in real-time, making the approach suitable for online trajectory planning in an AV using minimal computation resource. The 'windowed' nature of the approach means that the trajectory is being simultaneously planned for all sections of the road ahead, and thus it would be possible to adapt the technique for the rapid and efficient solution of dynamically changing road conditions - *e.g.* recovery after an unpredictable event, or adjustments in trajectory to avoid other road users (Erlien *et al.,* 2013). Though such a method would require extensive offline pre-training (*e.g.* using datasets from real or simulated road events), the parallel nature of the online planning task means that multiple paths could be predicted concurrently, thus the AV could be generating a trajectory through *all* possible routes in real-time - raising the exciting possibility of applications in overtaking manoeuvres, unexpected event recovery and accident-avoidance systems.

Due to the long computational times of many of the more accurate time-optimal approaches to finding the optimal line, a simple MCP or QSS pre-calculation is commonly employed as a starting point to expedite the process by constraining the problem (Kapania, 2016). The method presented in this paper could be coupled with a time-optimal simulator to provide an exceptionally fast, accurate initial approximation for use as pre-calculation prior to fine-tuning with further (*e.g.* minimum-time) optimisations.

# 5 Conclusions

This work has demonstrated the potential of machine learning techniques to generate an extremely rapid prediction of the optimal trajectory around a previously-unseen circuit. A feed-forward ANN was selected, with Sigmoid and Hard Sigmoid activation functions (on the hidden intermediate and output layers respectively) found to deliver the best performance.

This approach does not solve the problem *per se*, rather it rapidly generates an accurate prediction to the solution based upon prior training on many solutions to a similar problem - and therefore thousands of circuits were used to train the network. This study used track boundaries for 82 real circuits, which were expanded to 6058 tracks through various augmentations (including scaling / flipping / reversing *etc.*). An existing Optimal Control method was employed to generate a racing line for each track, and each circuit was split into sections via a sliding window approach - resulting in a dataset comprising a total of 2.7 million track segments.

The ANN will accept training data from any existing method of generating the racing line, offering a prediction based upon that data. Thus, if a large number of circuits driven by a human driver in a DIL simulator (or a complex and accurate method of calculating the line) were used - the ANN would 'learn to drive like a particular driver (or simulator) in a particular car'. Calculation time is independent of the model complexity used for training purposes, so if a highly-complex, free-trajectory simulator were used to generate the training data, the accuracy of the solution would increase without affecting prediction time.

Predictions of racing lines for previously-unseen circuits were found to be with an average ±0.27m mean absolute error, with greatest accuracy where it matters most - at the corner apex (±0.11m). Accuracy of the generated line is comparable to (in many cases better than) other traditional methods of obtaining the racing line. The prediction accuracy is comparable to a professional racing driver's ability to identify and follow the optimal line, and that of an autonomous vehicle control system's path-following accuracy.

The network provides the most accurate predictions where a breadth of similar features exist in the training data. In order to achieve the lowest prediction errors, the dataset should be centred around circuits of a similar size and complexity (*i.e.* if a prediction is desired for a small kart track, training the ANN on a dataset featuring only F1-style circuits would result in increased prediction error). Additionally, the training data should be diversified to reduce sample bias by using a large number of original circuits (rather than the large number of augmentations of a small number of circuits used in this study).

The ANN calculates a prediction for a full-size racing circuit in approximately 0.033 seconds, making this approach over 9,000 times faster than a rapid OCP method, and approximately 800 times faster than the most expedient non-trivial approach reported in the literature survey. However, it should be noted the ML approach requires extensive training on existing data, and many other solutions simultaneously generate a speed trace - which is currently beyond the scope of the method presented in this paper.

Extremely rapid prediction of an accurate racing line means that this approach is the fastest known method of planning a target trajectory for a complete (or partial) circuit. It is therefore ideally suited to a range of applications including; calculation of a target line for use in traditional (*e.g.* QSS) simulators; pre-calculation to reduce solution time of free-trajectory (*e.g.* OCP) simulators; real-time prediction of the racing line in an autonomous racing vehicle - and it could be adapted for use in overtaking manoeuvres or trajectory planning and accident avoidance for an autonomous road-going vehicle. This represents a re-think of the traditional approaches to obtaining an optimal trajectory - where previous approaches typically rely upon brute-force computation, this study suggests that for certain applications a data-driven approach may be considerably more efficient without any fundamental trade-off in accuracy.

## Acknowledgement

This project has been made possible with support from the Autonomous Driving & Intelligent Transport group at Oxford Brookes University, and the *OBR Autonomous* team. The authors would like to thank Peter Ball, Fabio Cuzzolin, Matthias Rolf, Alex Rast and Gordana Collier for their ongoing support. Later stages of this work were carried out in conjunction with *Manchester Stinger Motorsports FS-AI* at the University of Manchester, where credit goes to Caroline Jay, Robert Stevens and Paul Nutter for their support in founding the team. Special thanks go to the Institute of Automotive Technology at TU Munich - this work would not have been possible without their *Global Race Trajectory Optimisation Tool* (TUMFTM, 2019).